\documentclass[a4paper]{svproc}
\pdfoutput=1
\usepackage{url}
\usepackage{comment}
\usepackage{graphicx}
\usepackage{graphics}
\usepackage{amsmath}
\usepackage{multirow}
\usepackage{hyperref}
\usepackage{array}
\usepackage{tabularx}
\usepackage{soul}
\usepackage{color}
\usepackage[T1]{fontenc}
\usepackage{subcaption}
\usepackage[export]{adjustbox}

\newlength\subfiguresheight

\newcommand{\ph}[1]{{\textbf{#1:}}}
\newcommand{\rph}[1]{{\textbf{}}}

\definecolor{pink}{rgb}{0.98, 0.38, 0.5}
\newcommand{\chris}[1]{\textcolor{pink}{Chris: #1}}

 \newcommand{\submitted}{}

\newcommand{\preprint}{: Preprint Version}

\begin{document}
\mainmatter              
\title{BAXTER: Bi-modal Aerial-Terrestrial Hybrid Vehicle for Long-endurance Versatile Mobility\preprint}

\author{Hyungho Chris Choi\inst{1}
\and Inhwan Wee\inst{1}
\and Micah Corah\inst{2}
\and Sahand Sabet\inst{3}
\and Taeyeon Kim \inst{1}
\and Thomas Touma\inst{2}
\and David Hyunchul Shim \inst{1}
\and Ali-akbar Agha-mohammadi \inst{2}
}
\authorrunning{Hyungho Chris Choi et al.\preprint\submitted}
\titlerunning{BAXTER: Hybrid Aerial Vehicle\preprint\submitted}
%
%
\institute{
Korea Advanced Institute of Science and Technology, Republic of Korea\\
\and
Jet Propulsion Laboratory, California Institute of Technology, Pasadena, CA, USA
\and
Univesity of Arizona, Tucson, AZ, USA
}
\maketitle              
%
%
\begin{abstract}
\label{section:abstract}
%
Unmanned aerial vehicles are rapidly evolving within the field of robotics. 
However, their performance is often limited by payload capacity, operational time, and robustness to impact and collision.
These limitations of aerial vehicles become more acute for missions in challenging environments such as subterranean structures which may require extended autonomous operation in confined spaces.
%
While software solutions for aerial robots are developing rapidly, improvements to hardware are critical to applying advanced planners and algorithms in large and dangerous environments where the short range 
and high susceptibility to collisions of most modern aerial robots make applications in realistic subterranean missions infeasible.
%
To provide such hardware capabilities, one needs to design and implement a hardware solution that takes into the account the Size, Weight, and Power (SWaP) constraints.
%
This work focuses on providing a robust and versatile hybrid platform that improves payload capacity, operation time, endurance, and versatility.
%
The Bi-modal Aerial and Terrestrial hybrid vehicle (BAXTER) is a solution that provides two modes of operation, aerial and terrestrial.
BAXTER employs two novel hardware mechanisms: the \textit{M-Suspension} and the \textit{Decoupled Transmission} which together provide resilience during landing and crashes and efficient terrestrial operation.
%
%
Extensive flight tests were conducted to characterize the vehicle's capabilities, including robustness and endurance.
Additionally, we propose Agile Mode Transfer (AMT), a transition from aerial to terrestrial operation that seeks to minimize impulses during impact to the ground which is a quick and simple transition process that exploits BAXTER's resilience to impact.
%
\end{abstract}
\section{Introduction}
\label{section:01_Introduction}
\ph{Problem Statement and Challenges}
Consider an autonomous robot traversing through a large and complicated subterranean environment with narrow corridors, prone to impact, and disconnected stretches of ground-traversable terrain.
In such an environment, both conventional drones and unmanned ground vehicles (UGV's) would have limited scope of operation due to limited operational time and lack of traversability, respectively.
%
The problem mentioned above is a widespread challenge for traversing subterranean environments---including mines, caves, and urban sprawl---which require extensive endurance to impact and versatile solutions that can traverse across rough and steep terrain~\cite{about_subt}.
\begin{figure}[t!]
    \centering
    \includegraphics[width=\textwidth]{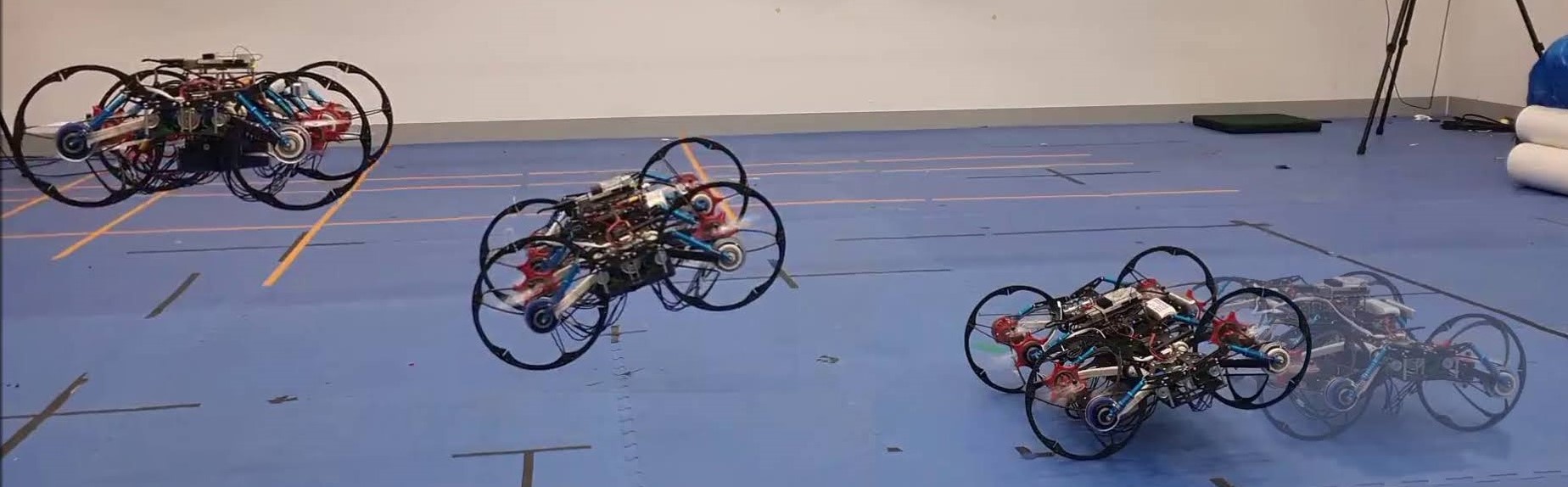}
    \caption{Agile Mode Transfer.
    BAXTER exhibits a simple transfer from aerial to terrestrial operation via a short free fall with body inclined to reduce impulses to wheels.}
    \label{fig:AMT_photo}
    \vspace{-.600cm}
\end{figure}
%
%
Although potential functions of drones continue to expand by various algorithmic advances \cite{davidfanrollocopter,thomaslewISRR}, the development of hardware platforms that simultaneously achieve both the versatility of aerial vehicles and the endurance of ground vehicles is an important topic for hardware engineering in robotics.
The challenges in developing such platforms include delicate management of payload and hardware features in limited size, weight, and power (SWaP) constraints, maintaining structural and aerial stability, and all-around robustness of co-dependent components (e.g. coupled drive and transmission ) to mitigate risk of system failure.

\ph{Proposed Work and Contribution}
 This publication outlines the development of the \textbf{B}i-modal \textbf{A}erial\textbf{-TER}restrial Hybrid Vehicle (BAXTER), shown in Fig.~\ref{fig:AMT_photo}, which aims to provide a robust hardware-based solution for the aforementioned challenges.
The highlights of the contributions are:
\begin{itemize}
	\item  \textbf{Improved Operating Scope and Range} via the efficient use and transition between two modes of operation: \textit{Terrestrial} (low-versatility and high-efficiency) and \textit{Aerial} (high-versatility and low-efficiency).
	\item \textbf{Improved Resilience} through introduction of wheel assemblies around the aerial propulsion units as well as the suspension system, emphasizing resilience to impact while reasonably considering SWaP constraints.
	\item \textbf{Novel Mechanical Concepts} (the \textit{M-Suspension} and \textit{Decoupled transmission})
	were introduced to realize the contributions above.
	\item \textbf{Agile Mode Transfer} demonstrates the versatility and resilience of the platform through a simple transition from aerial to terrestrial operation.
\end{itemize}

\ph{Outline} In this paper, novel mechanical concepts are proposed, realized, and tested to provide the above-mentioned contributions.
Section~\ref{section:02_Related_Work} summarizes notable developments in hybrid vehicles, morphing drone concepts, and general-purpose aerial vehicles.
Section~\ref{section:03:Problem_Statement} states the goals and context of this paper.
Section~\ref{section:04_Technical_Approach} explains the design concept and dynamics, and Sections~\ref{section:05_Experiments} and~\ref{section:06_Results} summarize the realization and testing of the first   prototype.
Sections~\ref{section:07_Experimental_Insights} and~\ref{section:08_Conclusion_and_Future_Works} provide insight into the development process, propose future works, and conclude.
%
%
\section{Related Work}
\label{section:02_Related_Work}
This section outlines previous hardware-based efforts on hybrid platforms, morphing drone concepts, and multi-functional aerial vehicles.
Additionally, the end of this section highlights the design iterations toward developing the BAXTER robot platform.

\ph{Hybrid Platforms} Hybrid Platforms are a class of platforms that include more than one mode of operation \cite{davidfanrollocopter,rollocopter,hytaq}. In the scope of this publication, unmanned hybrid platforms incorporating aerial and terrestrial operation are the main focus.
The current Hybrid platforms can be categorized in two ways:
\begin{itemize}
    \item \textbf{Separate Articulation in Terrestrial Operation}: Platforms utilizing separate means of propulsion in terrestrial mode are classified to have \textit{Active Terrestrial Mode} of operation. Vehicles with Active Terrestrial Mode \cite{drivocopter,fstar}, compared to vehicles with Passive Terrestrial Mode \cite{davidfanrollocopter,rollocopter,hytaq}, provide a more efficient and mutually independent operation, while their payload capacity is partially reduced due to the addition of the driving system.
    \item \textbf{Morphing Mechanism}: Platforms utilizing separate built-in  mechanisms for changing the internal form factor are \textit{Morphing} platforms. Platforms can be \textit{Active  Morphing} \cite{fstar}, \textit{Passive Morphing}, or \textit{Non-Morphing} \cite{davidfanrollocopter,rollocopter,hytaq}. Morphing capabilities in hybrid vehicles provide vehicles with specialized forms for each operation, but may limit payload capacity or scale. 
\end{itemize}
\begin{figure}[t!]
\centering
\setlength{\subfiguresheight}{0.22\linewidth}
\begin{subfigure}[b]{.28\linewidth}
    \centering
    \includegraphics[height=\subfiguresheight]{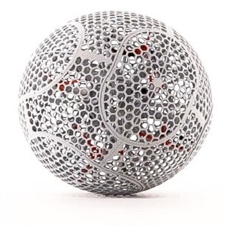} 
    \caption{Rollocopter \cite{rollocopter}}
    \label{fig:sub-first-rollo}
\end{subfigure}
\begin{subfigure}[b]{.40\linewidth}
    \centering
    \includegraphics[height=\subfiguresheight]{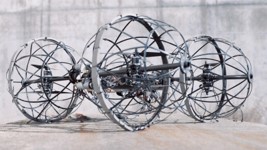}
    \caption{Drivocopter \cite{drivocopter}}
    \label{fig:sub-second-drivo}
\end{subfigure}
\begin{subfigure}[b]{.28\linewidth}
    \centering
  \includegraphics[height=\subfiguresheight]{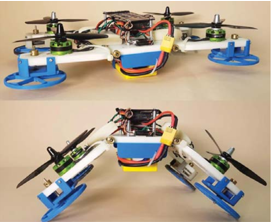}  
  \caption{Fstar \cite{fstar}}
  \label{fig:sub-third-fstar}
\end{subfigure}
\caption{Hybrid Vehicles and their Classifications.
(\subref{fig:sub-first-rollo}) Passive drive, non-morphing.
(\subref{fig:sub-second-drivo})  Active drive, non-morphing.
(\subref{fig:sub-third-fstar}) Active drive, active morphing.}
\label{fig:group}
\vspace{-.500cm}
\end{figure}
%

\textbf{Morphing Drone Concepts} 
Apart from hybrid platforms, this class of novel aerial vehicle design focuses on morphing or controlled dynamic reconfiguration of the aerial structure to increase efficiency or versatility.
Single-entity approaches manipulate the form of individual drones to achieve favorable characteristics. 
Examples include actuation of thrust vectors to achieve additional freedom in motion \cite{tiltdrone_zheng,voliro_kamel,tilt2_Ryll}, translation of the thrusters to assume narrow profiles for traversing tight spaces \cite{uzh_foldable_drone,agile_valentin,passive_morph_mueller}, along with other unique morphing concepts \cite{pufferbot_hedayati,deformable_zhao,passive_morph_mueller}.
%
Multi-entity approaches arrange or integrate many drone units with similar characteristics to achieve different forms and characteristics to produce multi-link chains \cite{reconfig_chain_huan,2d_multilink_zhao} or other topologies.

\textbf{All-around Aerial Vehicles}
General purpose aerial vehicles represent another class of aerial vehicle designs that focus on SWaP (Size, weight, and power) constrained exploration challenges.
Features of aerial vehicles specialized for exploration in confined spaces include protective cages \cite{fliyability_elios2} and reconfiguration for traversal of narrow passages \cite{prometheus_brown}.
Furthermore, the scope of aerial vehicle designs spans as far as extra-terrestrial environments \cite{marsheli_demo_balaram,marsheli_vision_bayard}.
Control systems for such newly developed vehicles are another topic of study to improve resilience \cite{impactresilient_liu} or efficiency \cite{dyn_ener_sabet}.

\subsection{Design Iteration Toward BAXTER}
\label{section:02-1_Design_Iteration}
\begin{figure}[t]
\begin{subfigure}{.32\textwidth}
  \centering
  \includegraphics[width=.95\linewidth]{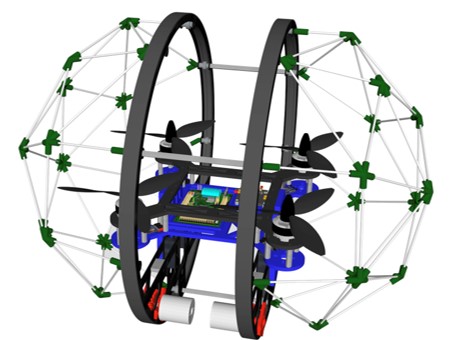}  
  \caption{MODEL 1 ``TUMBLER''} 
  \label{fig:sub-first-M1}
\end{subfigure}
\begin{subfigure}{.32\textwidth}
  \centering
  \includegraphics[width=.95\linewidth]{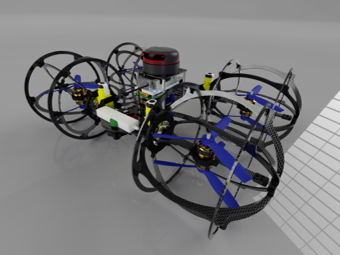}  
  \caption{ MODEL 2 ``HEXWING''} 
  \label{fig:sub-second-M2}
\end{subfigure}
\begin{subfigure}{.32\textwidth}
  \centering
  \includegraphics[width=.95\linewidth]{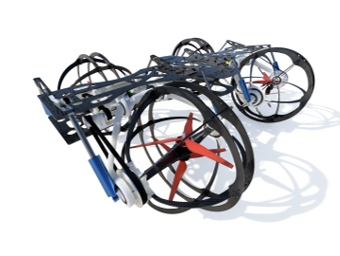}  
  \caption{MODEL 4 ``BAXTER''}
  \label{fig:sub-third-M4}
\end{subfigure}
\caption{Summary of Design Iterations.
(\subref{fig:sub-first-M1}) The initial working prototype of protected hybrid operation.
(\subref{fig:sub-second-M2}) The initial working prototype for spherical wheel-rotor subassembly.
(\subref{fig:sub-third-M4}) Incorporation of novel mechanical concepts to emphasize resilience and versatility.
}
\label{fig:fig}
\vspace{-.300cm}
\end{figure}

The development of BAXTER followed the design, manufacture, and evaluation of two predecessor models: MODEL 1 ``TUMBLER'' and MODEL 2 ``HEXWING''\footnote{One iteration, MODEL 3, was designed and not produced.} (illustrated in Fig.~\ref{fig:fig}).
Notable contributions of each design include:
\begin{itemize}
    \item \textbf{MODEL 1 ``TUMBLER''} An initial working prototype for a robot with a spherical protective cage and active terrestrial operation. Limited in payload capacity and versatility (Fig. \ref{fig:sub-first-M1}).
    \item \textbf{MODEL 2 ``HEXWING''} A proof of concept with spherical subassemblies around each arm. Increased versatility in terrestrial operation. Limited in resilience, scalability, and payload capacity. (Fig. \ref{fig:sub-second-M2})
    \item \textbf{MODEL 4 ``BAXTER''} Incorporation of novel mechanical concepts (the \textit{M-Suspension} and the \textit{Decoupled Transmission}) to emphasize resilience and versatility. Increased payload capacity. (Fig. \ref{fig:sub-third-M4})
\end{itemize}


%
%
%

%
\section{Problem Statement}
\label{place:problem_statement}
\label{section:03:Problem_Statement}
\rph{Section outline}
This section describes the objectives for the development of BAXTER with appropriate context.

\ph{Goal statement}
The development of BAXTER aims to provide a hardware-based solution for the following capabilities:
(1) A hybrid vehicle that significantly increases operational scope and range by utilizing two modes of travel: \textit{Aerial} and \textit{Terrestrial}.
(2) A collision resistant hardware platform to maintain each mode and transition between them.
(3) A robust, non-specific hardware platform for use in general, third party applications.
(4) A controlled demonstration of the transition from aerial to terrestrial mode.
%
Due to energy expenses in flight, the main mode of operation for BAXTER is Terrestrial, and its aerial operation is intended to be briefly used to traverse through environments that do not permit terrestrial operation.
This addresses navigation through challenging subterranean environments such as tunnels, subway stations, and caves, which include tight and narrow spaces where many existing platforms would be prone to collision.
Further, the design directives described above are in line with the challenging environments from the DARPA Subterranean Challenge, which inspired the development of BAXTER \cite{about_subt}.
%
%

%
\section{Technical Approach}
\label{place:mechdesign}
\label{section:04_Technical_Approach}
\rph{Section Outline} 
This section describes the BAXTER hardware in detail along with the simple dynamic basis for the demonstrative landing procedure.

\begin{figure} [t!]
    \includegraphics[width=\textwidth]{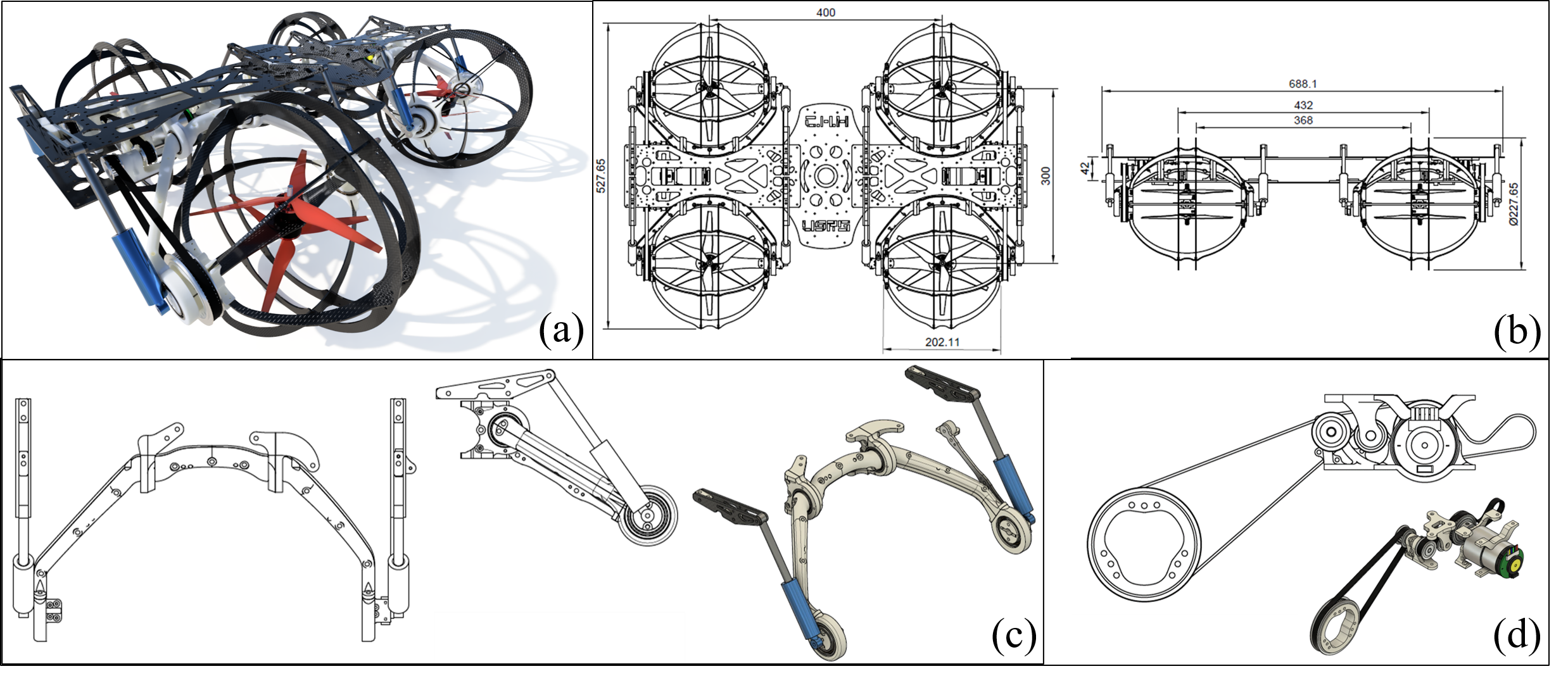}
    \caption{ (a) Rendering of BAXTER's Full Body Without Sensors/Processors (b) Overall dimensions of the platform (c) The M-Suspension design provides resilience during landing, crashes, and terrestrial operation (d) The Decoupled Transmission delivers motor-power to each wheel through a moving shock absorbing system.}
    \label{fig:cad_summary}
    \vspace{-0.300cm}
\end{figure}

\textbf{Base Design}: Toward realizing the capabilities discussed in Section~\ref{place:problem_statement}, BAXTER's design falls into the class of robots with an \textit{active terrestrial mode} and \textit{passive morphing}, according to Section~\ref{section:02_Related_Work}.
The design is centered on a concept of rotating, separately powered cages around each unit of propulsion (Fig.~\ref{fig:cad_summary}).
For aerial operation, an X8 configuration\footnote{The X8 configuration is a planar configuration of 8 rotors, arranged in 4 coaxial pairs.} is used with 7-inch rotors mounted on 2500KV class motors.
For terrestrial operation, a single DC motor powers each pair of wheels via the novel transmission system.
The rendering and overall dimensions of BAXTER's mechanical design are illustrated in Fig.~\ref{fig:cad_summary} (a)-(b).

\ph{Novel Concepts} BAXTER features novel mechanical suspension and motor-power transmission systems.
The need for reliable shock resistance for both drop impact and terrestrial operation motivated the incorporation of a suspension system into the aerial platform.
The \textit{M-Suspension} (Fig.~\ref{fig:cad_summary}-(c)) is a novel approach for vertical shock reduction in a hybrid aerial vehicle.
This new type of passive suspension design focuses on ease of manufacturing based on 3D printing and minimizing the amount of  additional weight.
However, this suspension introduces a new obstacle: transmitting power through the moving joint during active terrestrial operation.
The \textit{Decoupled Transmission} (Fig.~\ref{fig:cad_summary}-(d)) is a power transmission that has a timing pulley on an axis coinciding with each suspension joint.
This design keeps the total pitch line length of the timing belt constant at all angles of the suspension joint (so that the belt tension remains constant at all times).
In conjunction with the suspension design, this completes the mechanical concept to deliver motor power to each wheel through a moving, shock absorbing system.

%
%
%

\subsection{Agile Mode Transfer}
\label{sec:amt}
A simplified kinematic model of a quick and simple procedure for transfer from aerial to terrestrial mode referred to as \textit{Agile Mode Transfer (AMT)} is discussed here (see Fig.~\ref{fig:AMT_photo}).
%
AMT includes a controlled impact with the ground at an angle which provides a fast transition from aerial to terrestrial operation.

The basis of the simplified kinematics of AMT is outlined below.
Figure~\ref{fig:physics} illustrates the four phases of the process
(Free Fall, Initial Impact, Roll, and Final Impact). The subscripts used in Fig.~\ref{fig:physics} indicate
the number of the phase associated with given variable.
In the following, 
$\Vec{v}_n$ and $\vec{\omega}_n$ respectively represent the translational velocity and the angular velocity of the center of mass (CoM) of the chassis at the end of the respective phase.
It is assumed that the CoM of the system is located at the center of the chassis. The term $\vec{h}_n$ represents the final height of the  CoM, vector $\vec{d}_2$ connects the contact point  of the \textit{Back Wheel} to the CoM, and vector $\vec{d}_4$ connects the contact point  of the \textit{Front Wheel} to the CoM. The four phases shown in Fig.~\ref{fig:physics} are discussed below:  
\begin{figure}
    \centering
    \includegraphics[width=\textwidth]{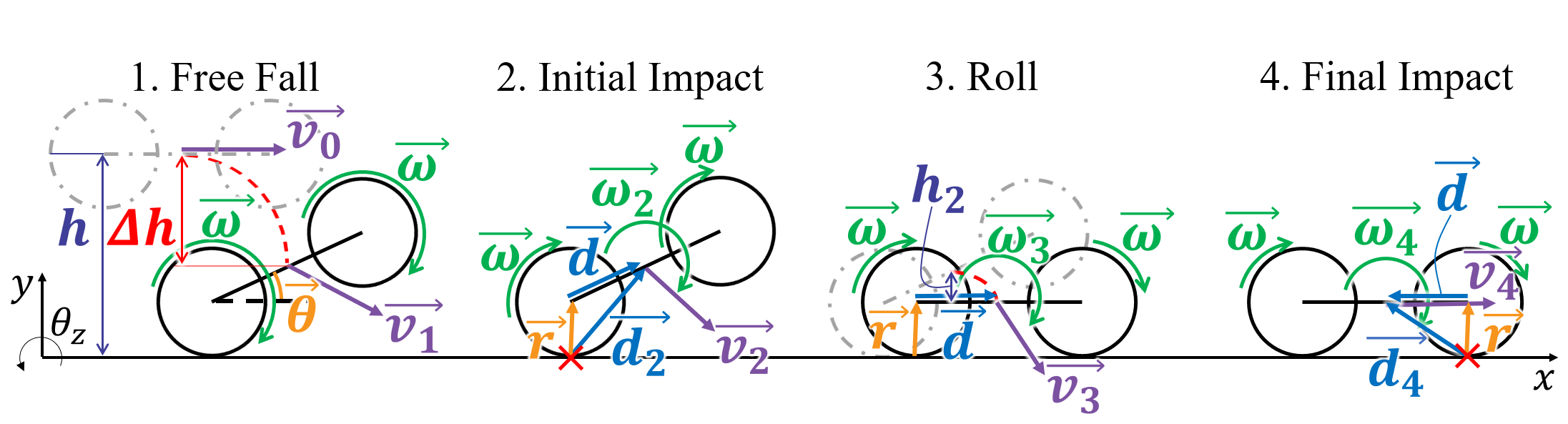}
    \caption{Simplified Kinematics Model for BAXTER's Agile Mode Transfer (AMT). AMT models the landing procedure as four phases:
    1. Free Fall, 2. Initial Impact, 3. Roll, 4. Final Impact.
    }
    \label{fig:physics}
    \vspace{-.600cm}
\end{figure}

\begin{enumerate}
    \item \textbf{Free Fall}: Vehicle keeps a positive landing angle ($\theta$) and is subjected to free-fall motion until making contact with the ground. Governing principle: parabolic motion.
    \begin{align}
         v_{1x} =& v_{0x} &
         v_{1y} =& -\sqrt{2g\Delta h}
    \end{align}
    Here $\Delta h$ is the magnitude of the vertical displacement of the CoM during free fall, $v_{0x}$ is the desired entrance velocity in x-direction, and $g$ is the gravitational acceleration.
    \item \textbf{Initial Impact}: The first wheel (i.e. \textit{Back Wheel}) makes contact with the ground. Governing principles: Conservation of angular momentum and no-slip condition.
    \begin{align}
         &\Vec{d}_2 \times \Vec{v}_1 = \Vec{d}_2 \times \Vec{v}_2 + I\Vec{\omega}_2 &
         & \Vec{\omega}_2\times\Vec{d} + \Vec{\omega} \times \Vec{r}= \Vec{v}_2 
    \end{align}
    Here $\vec{d}$ is the displacement vector connecting the \textit{Back Wheel} center to the CoM and $\vec{r}$ connects the contact point to the center of the \textit{Back Wheel}.
    \item \textbf{Roll}: The first wheel rolls while the second wheel is in motion toward the ground. Governing principles: conservation of energy and no-slip condition.
    \begin{align}
         \frac{1}{2}mv_2^2+\frac{1}{2}Iw_2^2+mgh_2 &= \frac{1}{2}mv_3^2+\frac{1}{2}Iw_3^2&\\
          \Vec{\omega_3}\times\Vec{d} + \Vec{\omega} \times \Vec{r} &= \Vec{v_3}&
    \end{align}
    Here the terms $\Vec{\omega}$, $m$, and $I$ present the angular velocity, mass, and moment of inertia of the chassis, respectively.
    \item \textbf{Final Impact}: The second wheel (i.e. \textit{Front Wheel}) makes contact with the ground. Governing principles:  Conservation of angular momentum, no-slip condition for both wheels.
     \begin{align}
     \Vec{d}_4 \times \Vec{v}_3 + I\Vec{\omega}_3 &=
     \Vec{d}_4 \times \Vec{v}_4 + I\Vec{\omega}_4 &
     \Vec{\omega}\times\Vec{r} + \Vec{\omega}_4\times\Vec{d} &= \Vec{v}_4 &
     \vec{v_4} &= \vec{v_f}
    \end{align}
\end{enumerate}
\par Given the desired entrance and exit velocity ($v_0, v_f$), this model is used to determine the optimal landing angle ($\theta$) that minimizes the maximum impact with the ground. To achieve this angle, the impact impulses at phase 2 ($\Vec{I}_2$) and phase 4 ($\Vec{I}_4$) are defined as:
\begin{equation}
    \begin{aligned}[c]
        \Vec{I_2} = m\Vec{v}_2 - m\Vec{v}_1
    \end{aligned}
    \qquad
    \begin{aligned}[c]
        \Vec{I}_4 = m\Vec{v}_4 - m\Vec{v}_3
    \end{aligned}
\end{equation}
Then, the maximum impact $I_{\max}$ is defined as:
\begin{equation} \label{equation:optimum_criteria}
    I_{\max} = \max(|\Vec{I}_2|,|\Vec{I}_4|)
\end{equation}
Finally, the landing angle ($\theta$) that minimizes $I_{\max}$ is calculated numerically.
\section{Experiment Design}
\label{section:05_Experiments}
\rph{Section Outline}
Preliminary flight tests with the prototype verified BAXTER's viability.
Now, in this section, additional tests (outlined below) verify the flight characteristics, versatility, and resilience of the platform.

\ph{Basic Flight and Endurance Tests} Basic maneuvers (hover, rectangle, etc.) validate the flight characteristics of the prototype.
%
The results from these tests provide a quantitative evaluation (\textit{Flight and Mission Time, Maximum Speed, and Control Limits}) of the vehicle along with qualitative assessments.

\ph{Drop Test} BAXTER's suspension system is unique since it is designed to absorb step-like impulses from impact with the ground on landing or crashing.
So, to measure the suspension endurance, a drop test was performed with the prototype in the all-up configuration\footnote{All-up refers to the maximum payload configuration which uses additional weights as a proxy for sensing and computation payloads.} (4.2 kg total) and in an upright position from incrementally increasing heights to obtain the following metrics (Drop Limits):
(1)\textit{Maximum Intact Height}, the maximum height where the prototype takes no irreversible damage, and
(2) \textit{Maximum Flyable Height}, the maximum height where the prototype can still maintain aerial operation.

\begin{figure}[t!!!!]
    \centering
    \includegraphics[width=\textwidth]{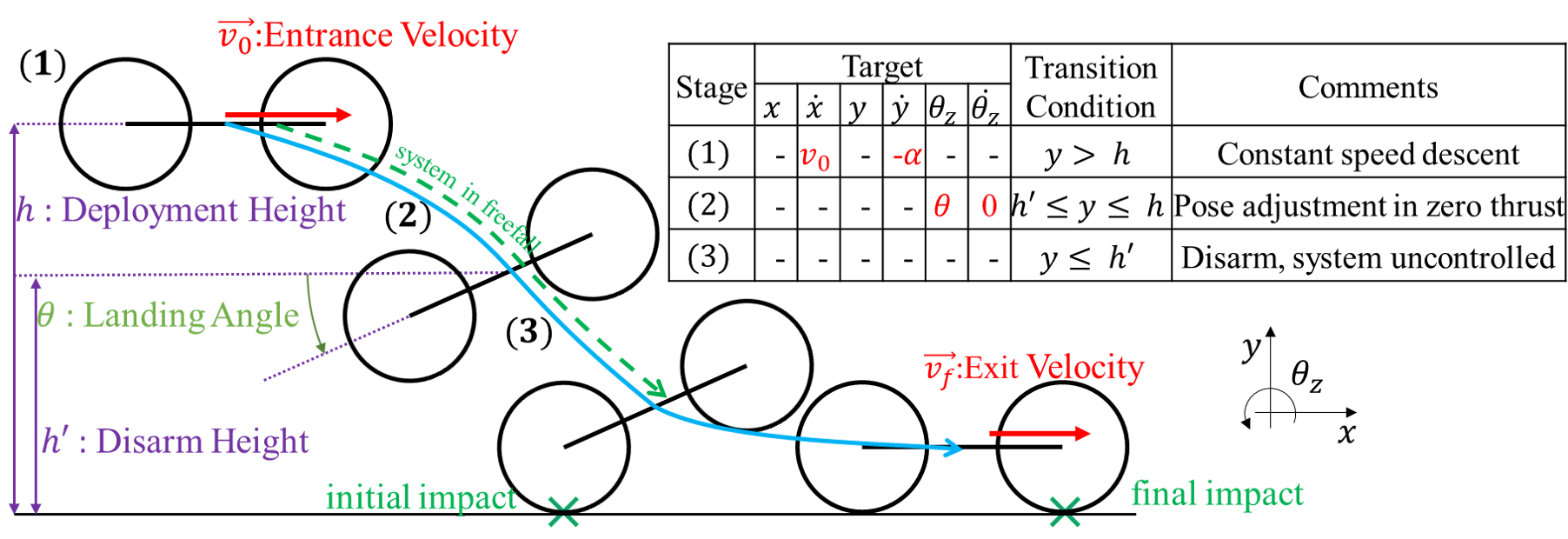}
    \caption{Agile Mode Transfer Mission Diagram, Mission Controller Stages (which are distinct from the phases of the landing process in Section~\ref{section:04_Technical_Approach}) and Parameters. This figure and the attached target table (controlled set-points for each phase are highlighted in red) illustrates and describes targets for the AMP in the mission perspective with 3 Stages:
    (1): Before Deployment,
    (2): Deployment (vehicle adjusts pose in free fall), and
    (3): Disarm (flight controller switches off)
    }
    \label{fig:land_test}
    \vspace{-.300cm}
\end{figure}

\begin{figure}[t]
    \centering
    \includegraphics[width=\textwidth]{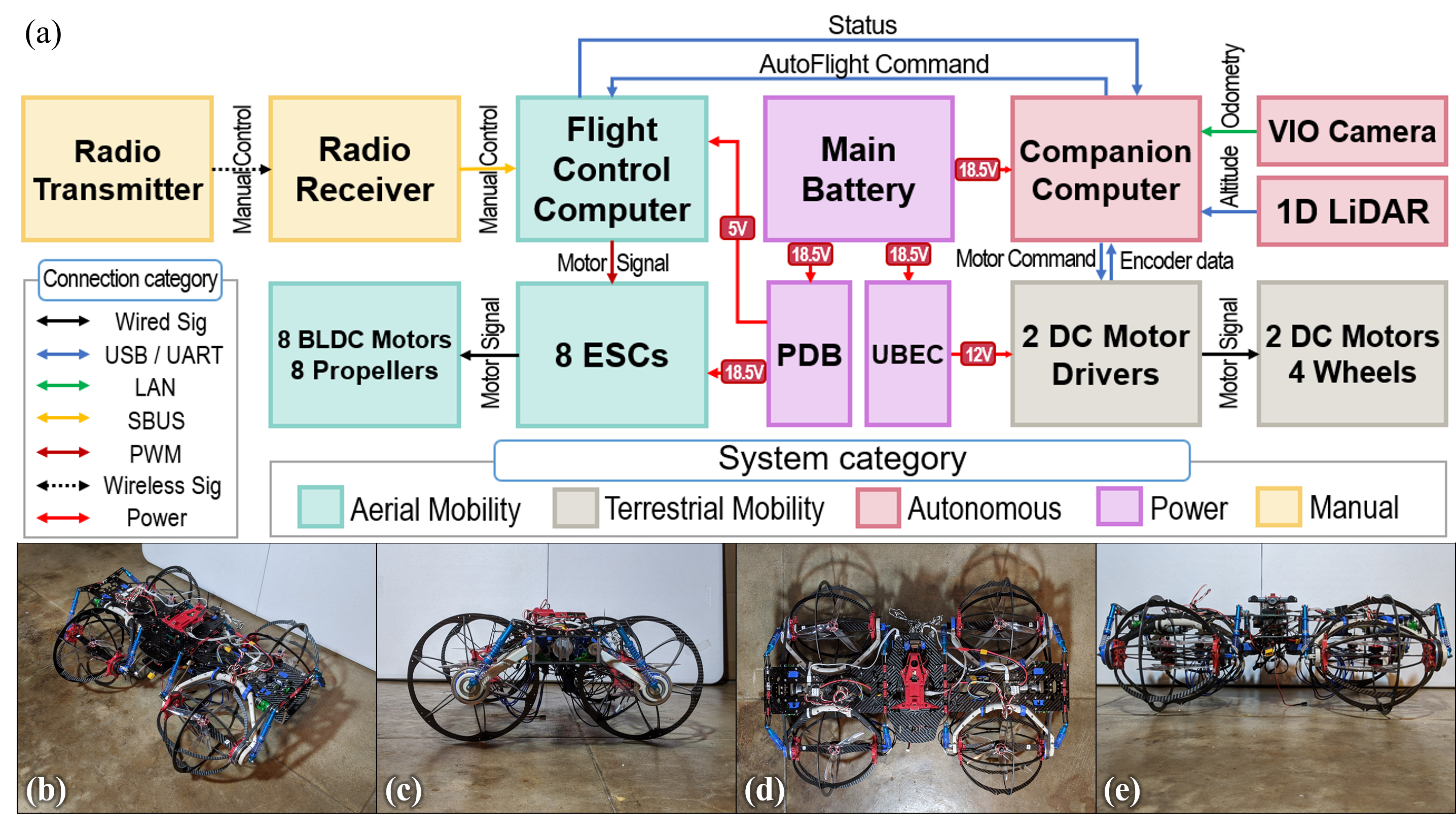}
    \caption{Overview of the First Prototype of BAXTER. (a) Electrical System Diagram of BAXTER (b)-(e) Aspect Photos of BAXTER Prototype
    }
    \label{fig:components}
 \vspace{-.600cm}
\end{figure}

\ph{Agile Mode Transfer Test} The agile mode transfer  experiments exploit and demonstrate the effectiveness of the hardware design.
From the kinematics model described in Section~\ref{section:04_Technical_Approach}, a mobile landing procedure will be devised by providing the parameters in Fig.~\ref{fig:land_test}, and calculating the optimum landing angle ($\theta$) that minimizes the load at impact in landing ($I_{max}$).
%
The implementation of this maneuver is accomplished via ROS and the attitude and velocity control capability of the flight controller \cite{ros_pixhawk_control}, with input from onboard sensors (Fig.~\ref{fig:components}).

\ph{Autonomous Flight System} To operate a mission autonomously in terrestrial and aerial mode, BAXTER uses odometry for velocity and attitude.
Attitude estimation is done through the 9-axis IMU (Inertial Measurement Unit) within the flight controller.
For velocity and position estimation, an Intel RealSense T265 tracking camera was  linked to a PX4 via ROS (Robot Operating System) running in the companion computer through VIO (Vision Inertial Odometry).
%
The mission controller for AMT runs on the companion computer (Nvidia Jetson TX2), which implements the 3 mission phases outlined in Fig.~\ref{fig:land_test}.
%
%
%

\ph{Summary of Prototype} Finally, the first prototype of BAXTER presents a tangible proof of concept for the mechanical design described in Section~\ref{place:mechdesign}.
The details of the circuit-level hardware are listed in Fig.~\ref{fig:components}-(a).

\begin{table}[t!!!!!!]
    \centering
    \caption{Obtained Flight Parameters for BAXTER. (\protect\subref{table:result_measurement}) Obtained results from measurements
    (\protect\subref{table:result_experiment}) Obtained results from experiments
    }\label{table:results}
    \scalebox{0.9}{
      \begin{subfigure}[t]{0.5\linewidth}
        \caption{}\label{table:result_measurement}
        \begin{tabular}{c||c c}
            \hline
            \textbf{Category}&\textbf{Sub-Category}&\textbf{Metric}\\
            \hline
            \textbf{Mass}
            &Chassis&1.2 kg\\
            &Propulsion System&1.2 kg\\
            &Maximum Payload&1.8 kg\\
            \hline
            \textbf{Inertia}
            &$I_{xx}$(roll axis)&0.351 $kg \cdot m^2$\\
            \textbf{ (All-Up)\footnotemark[6]}
            &$I_{zz}$(pitch axis)&0.125 $kg \cdot m^2$\\
            &$I_{yy}$(yaw axis)&0.254 $kg \cdot m^2$\\
            \hline
        \end{tabular}
      \end{subfigure}%
      \hspace{4em}%
      \begin{subfigure}[t]{0.5\linewidth}
        \caption{}\label{table:result_experiment}
        \begin{tabular}{c||c c}
            \hline
            \textbf{Category}&\textbf{Sub-Category}&\textbf{Metric}\\
            \hline
            \textbf{Time}
            &Flight (Manual)&5 min\\
            \textbf{Limit}
            &Flight (Autonomous)&4 min\\
            &Drive (Drive Only)&30 min\\
            \hline
            \textbf{Control}
            &Aerial Speed&2 m/s\\
            \textbf{Limit}
            &Terrestrial Speed&1 m/s\\
            &Roll Angle&30 deg\\
            &Pitch Angle&25 deg\\
            \hline
            \textbf{Drop}
            &Max. Intact&1.3 m\\
            \textbf{Limit}
            &Max. Flyable&1.4 m\\
            \hline
        \end{tabular}
      \end{subfigure}
    }
     \vspace{-.600cm}
\end{table}

\section{Results}
\label{section:06_Results}
\rph{Section Outline}
This section provides the qualitative and quantitative outcomes of the design and testing of BAXTER, described in Section~\ref{section:05_Experiments}.

\ph{Basic Flight Parameters}
Table~\ref{table:results} presents the results of the \textit{Basic Flight and Endurance Tests} and \textit{Drop Test} outlined in Section~\ref{section:05_Experiments}. These results were obtained and verified over 3 trials. Inertial measurements were conducted with the bifilar pendulum method \cite{bifilar_method}.

\begin{figure}[t]
    \centering
    \includegraphics[width=\textwidth]{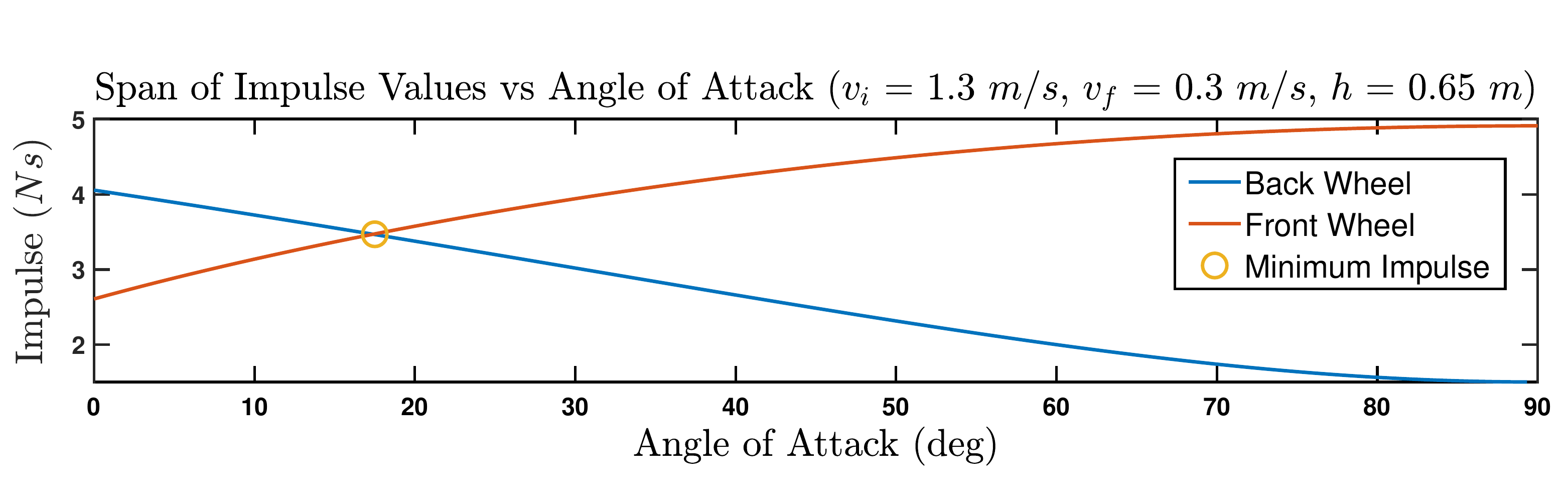}
    \caption{Plot of expected front ($I_2$) and back ($I_1$) wheel impulse values, varying the landing angle ($v_i = 1.3 m/s, v_f = 0.5 m/s, h = 0.65 m$) using the kinematics model described in Section~\ref{sec:amt}.
    The optimum landing angle, yielding the lowest maximum impulse value, is highlighted with a yellow circle.
    }
    \label{fig:AMT_matlab}
     \vspace{-.600cm}
\end{figure}

\ph{Agile Mode Transfer Test}
With the verified maximum intact drop height of around 1.3 m and control limits drawn from the previous experimental results, agile mode transfer solutions (varying Entrance and Exit Velocity) that were within the control limits were tested and verified to be repeatable.
Figure~\ref{fig:AMT_matlab} depicts the case of: $v_i = 1.3, v_f = 0.5, h = 0.65$ with respect to potential landing angles ($\theta$).
It can be deduced that the landing angle of around 20 deg produces the least maximum impulse on impact.
Such optimal landing angles are calculated internally to the mission controller and are implemented as a sequence of controller stages described in Fig.~\ref{fig:land_test}.
A representative frame-on-frame picture of this landing procedure is shown in Fig.~\ref{fig:AMT_photo}.
\section{Experimental Insights}
\label{section:07_Experimental_Insights}
\rph{Section Outline}
This section lists notable insights obtained regarding the overall flight characteristics and test results of BAXTER.

\ph{Overall Insight} Much insight into suspension systems for hybrid platforms was gained through development and experimentation on BAXTER.
To the best of our knowledge, BAXTER is the first platform to incorporate a passive suspension system into a hybrid aerial vehicle.
Notably, BAXTER's flight characteristics are affected by deflection of the suspension joints, but fine adjustment of the flight parameters allowed for reasonable performance in autonomous control.

\ph{Vibration Due to the Suspension System} This was evident in low-frequency vibration of the vehicle (magnitude of $1 Hz$) during preliminary flight tests.
Although these vibrations can be mitigated by careful tuning of the flight controller, doing so restricts the control limits.

\ph{Motor Translation due to Body Distortion} As a side effect of deflecting passive suspension joints, passive morphing occurs, and the thrusters translate along with the wheel assemblies in flight.
This introduces irregularities in flight behavior due to distortions of the body that are not present in conventional aerial vehicles.
A variant of this phenomenon has been studied with respect to a folding drone where a motor-actuated morphing mechanism involving translating thrusters introduced undesired displacement in flight behaviour \cite{uzh_foldable_drone}.
A potential research topic can be to dynamically change parameters in the control system to mitigate the deflection of the suspension system in order to expand the control limits for BAXTER.
%

\ph{Insight on Resilience}
As intended, the results from the drop tests (see Section~\ref{section:05_Experiments}) demonstrate the resilience of BAXTER in contact-laden environments and crash-prone operation.
A notable takeaway from the drop tests is that for many times that the prototype took damage its onboard electrical components remained intact while primarily easily-replaceable 3D-printed parts took concentrated damage.
A potential future design objective cloud be to concentrate structural weak spots within the platform to easily accessible parts (Commercially available parts, 3D printed parts, etc.).

\ph{Insight on Agile Mode Transfer}
Agile Mode Transfer tests were a successful demonstration of BAXTER's potential for improved operational scope and resilience.
Overall, BAXTER was able to reliably perform Agile Mode Transfer (AMT) without irreversible damage to the hardware.
\section{Conclusion and Future Works}
\label{section:08_Conclusion_and_Future_Works}
\ph{Summary of BAXTER}
BAXTER presents a novel hardware-based approach to aerial-terrestrial hybrid platforms.
It also offers a solution for multiple fronts in challenging environments where both resilience and extensive operational scope and range are required.
This publication has detailed the design philosophy and prototype of the BAXTER concept and successfully showed viability of its hybrid capabilities by evaluation of conventional flight characteristics and testing the novel Agile Mode Transfer.

\ph{Future Works}
BAXTER provides a reliable platform to operate in large and contact-prone environments where operation of conventional aerial robots would prove to be infeasible.
BAXTER also opens avenues for applications (such as subterranean exploration) that exploit its resilience which is demonstrated with the Agile Mode Transfer.
%
Introducing a contact-flexible platform also opens potential for new software features that exploit contacts such as for computation of odometry~\cite{thomaslewISRR}
as well as risk-aware mapping~\cite{agha2019confidence} and planning~\cite{agha2018slap},\cite{Agha17IROS}
.



%
%
Further iteration on this design will continue to push the boundaries of hybrid robot  platforms.
For BAXTER, the introduction of the \textit{M-Suspension} and the \textit{Decoupled Transmission} provided notable improvements.
But, as described in Section~\ref{section:07_Experimental_Insights}, challenges in aerial mobility arise as a trade-off between including a passive suspension system for resilience and adverse changes to the vehicle's flight characteristics.
One hardware-based solution to this challenge is the introduction of a simple active suspension, specifically a suspension system that can be switched on or off.
This change to the suspension system will be a main focus for the design of the next iteration (MODEL 5).

%

%
\section*{Acknowledgements}
This work was supported by the Institute for Information \& communications Technology Promotion (IITP), funded by the Korean government (MSIP) (Development of AI-powered Autonomous Drone for Complex Indoor Environment) under the guidance of Unmanned System Research Group of Korea Advanced Institute of Science and Technology (KAIST USRG).
Further guidance was provided from the Jet Propulsion Laboratory, California Institute of Technology, under a contract with the National Aeronautics and Space Administration (80NM0018D0004).
We want to thank Hyunjee Ryu, Brian Kim, and Hanseob Lee of KAIST USRG, along with Brett Lopez and Team CoSTAR of JPL for various technical support and fruitful discussions.
\footnotesize{
\bibliographystyle{splncs}
\bibliography{99_references}
}
\end{document}